This article has been accepted for publication in IEEE Transactions on Cognitive and Developmental Systems. This is the author's version which has not been fully edited and content may change prior to final publication. Citation information: DOI 10.1109/TCDS.2022.3216911

# GRIMGEP: Learning Progress for Robust Goal Sampling in Visual Deep Reinforcement Learning

Grgur Kovač, Adrien Laversanne-Finot, Pierre-Yves Oudeyer

***Abstract*—Autotelic RL agents sample their own goals, and try to reach them. They often prioritize goal sampling according to some intrinsic reward, ex. novelty or absolute learning progress (ALP). Novelty-based approaches work robustly in unsupervised image-based environments when there are no distractors. However, they construct simple curricula that don't take the agent's performance into account: in complex environments, they often get attracted by impossible tasks. ALP-based approaches, which are often combined with a clustering mechanism, construct complex curricula tuned to the agent's current capabilities. Such curricula sample goals on which the agent is currently learning the most, and do not get attracted by impossible tasks. However, ALP approaches haven't so far been applied to DRL agents perceiving complex environments directly in the image space. GRIMGEP, without using any expert knowledge, combines the ALP clustering approaches with novelty-based approaches and extends them to those complex scenarios. We experiment on a rich 3D image-based environment with distractors using novelty-based exploration approaches: Skewfit, and CountBased. We show that wrapping them with GRIMGEP - using them only in the cluster sampled by ALP - creates a better curriculum. The wrapped approaches are attracted less by the distractors, and achieve drastically better performances.**

## I. INTRODUCTION

RECENT work in reinforcement learning (RL) has shown that robots can learn complex individual skills such as grasping [1], locomotion [2], [3], and manipulation tasks [4]. However, the standard paradigm in reinforcement learning often involves learning policies specific to each task. This imposes an engineering burden as a reward function has to be manually designed for each task. On the contrary, humans are capable to autonomously learn a wide variety of skills by setting their own goal and practicing them, with little to no supervision. They naturally move from simple tasks in the early infancy to more complex tasks as they grow. Designing autonomous learning agents, capable of setting their own goals, similarly to how humans learn is thus a promising avenue of research. Such agents would autonomously discover general purpose policies, that could later be used for specific purpose, reducing the need for outside intervention and manual engineering.

In the absence of a specific objective, an efficient strategy in order to learn a general purpose policy is to learn to reach as many states as possible [5]–[9] (see [10] for a review). Combining a goal reaching policy with an exploration mechanism, even in its simplest form, can allow solving otherwise impossible environments. For example, GO-Explore [11] showed that when goal-reaching is assumed solved by expert knowledge (manually setting the environment state to the goal state), performing random actions from interesting goals states allows to solve hard exploration problems.

This form of goal exploration involves two challenges: how to learn a goal-reaching policy and how to efficiently explore by choosing the goals to be reached.

This first challenge is solved by using recent reinforcement learning algorithms [12], [13] together with good goal representations. This goal representation can be either engineered from the state space [9], [14] or learned by the agent during exploration [5], [8], [15].

The second issue relates to the problem of constructing an automatic curriculum. An ideal curriculum should choose the task on which the agent will learn the most, gradually moving from easier tasks to more complex ones [16]–[19]. An efficient curriculum will lead the agent to favor tasks for which the agent is making progress or tasks that are starting to be forgotten, over tasks which have already been learned, are too hard at the moment, or are unlearnable. In other words, this curriculum should adapt to the current capabilities of the agent.

In order to solve this second issue, many approaches try to maximize the diversity of sampled goals, for example by using importance sampling techniques [6] in order to sample states uniformly from the underlying distribution. While optimizing the diversity of sampled goals has been shown to work well in high dimensional image state spaces [6], it falls short in more complex environments requiring a more complex automatic curriculum. One example of such a problem are environments containing unpredictable distractors. This problem is often referred to as the noisy TV problem, as an agent aiming at maximizing novelty would get attracted by a TV displaying a sequence of continually novel and unpredictable images [20], [21].

An alternative way to construct a curriculum is to prioritize regions of the environment via learning progress estimates [17], [22], also referred as competence progress when it measures progress towards self-generated goals [23], [19]. For example, SAGG-RIAC [23] gradually clusters the environment into subregions of different learning progress and samples goals in the more promising regions (e.g. the regions with high learning progress). This approach has been shown to enable learning of inverse models in high-dimensional robots. In other approaches, the goal space is organized along several modules encoding for different types of goals: in that case, goal sampling is hierarchical [19], [9]. Learning progress has been shown to enable agents to autonomously build an efficient curriculum, and to help agents avoid unlearnable tasks [19] [9]. However, using learning progress estimates to sample goals in high dimensional image based environments is challenging due to the high dimensionality of the observations. As a result, goal-exploration experiments in the image-based setting has only been done in simple environments, with population based agents [8].

In this paper, we want to merge these two research directions: novelty-based exploration (which is robust in the image space) with ALP-based estimates (which are able to construct complex curricula). We provide a solution to extend ALP based approaches to image-based environments for RL-based goal exploration algorithms, and to combine them with novelty seeking exploration approaches.

The general idea is shown in figure 1. The environment is clustered into different parts (regions) for which the agent can then estimate the associated learning progress. An interesting region is selected based on its learning progress estimate, and a novelty estimate is used to select a goal from the selected region. From a high level perspective, our method can be viewed as a way to learn a prior over possible goals. The purpose of this prior is to detect which regions of the environment should be explored. This prior can then be combined with any goal exploration algorithm in order to guide exploration. We experimentally demonstrate that our framework improves the overall performances of various novelty-based goal exploration algorithms and allow the agent to autonomously focuses the exploration on interesting and learnable regions of the environment.

The main contributions are:

- A proof of concept: goal-based exploration can be extended to image-based unsupervised scenarios by combining, though clustering, ALP-based exploration on a higher level with novelty-based exploration on a lower lever.
- We introduce the GRIMGEP [1] framework: it extends goal sampling approaches based on clustering and learning progress to visual deep reinforcement learning agents.
- We combine it with existing novelty-based goal exploration algorithms, and show that it improves the performances and robustness of these algorithms.
- We provide an easily customizable 3D image-based environment that allows to study goal-based exploration in the presence of noisy distractors.

The paper is organized as follows. First, in related work, similar research on goal exploration, automatic curriculum, intrinsic motivation, and ALP is discussed. Then, the problem of goal exploration is defined. In section IV, two novelty-based goal exploration algorithms are explained: Skewfit, and CountBased. Those two algorithms will later be combined with GRIMGEP. In section V the GRIMGEP is introduced. It is explained first at a high level, then each component separately, and, lastly, the implementation details are provided. In the experiments section, the Explore3D environment is presented. Then, experiments are shown studying the behaviour of novelty-based approaches alone and those same approaches wrapped with the GRIMGEP framework. Finally, an ablation study of the effect of ALP is shown. The paper ends with a conclusion and a discussion of limitations and future work.

[1]GRIMGEP stands for **G**oal **R**egions guided **I**ntrinsically **M**otivated **G**oal **E**xploration **P**rocess.

For visualizations and the source code see https://sites.google.com/view/grimgep.

## II. Related work

**Goal exploration algorithms** Intrinsically Motivated Goal Exploration Processes (IMGEPs) are a class of exploration algorithms that explore by repeatedly setting goals for themselves, which they then try to achieve. Combined with an efficient curiosity mechanism for sampling goals, this approach has been shown to enable high-dimensional robots to learn very efficiently locomotion [24], manipulation of objects [6], [25], [26], navigation [27], or tool use [19]. Methods using absolute learning progress [23] to drive goal sampling were shown to scale up to real world environments with many forms of distractors, including action-induced distractors [9], [19]. These approaches often separate the goal space into regions and then compute ALP for each region. These regions are often defined manually [9], [19]. In order to reduce the burden of handcrafting these regions, methods have been designed to automatically separate the goal space into different regions using a clustering mechanism [23], [28]. However, these works relied on abstract hand-defined goal and state spaces (e.g. based on object positions and velocities). An exception is [8], which used learning progress to sample goals in a learned latent space, but this work studied population-based learners and required an offline pre-training. In this work, we adapt the learning progress approach to Deep RL goal-exploration algorithms that perceive their environments through pixels.

Recently, some approaches studied goal exploration in the image-based goal-conditioned Deep RL framework. For instance, [5] learns a goal-conditioned policy on top of a learned embedding of the environment. [7] proposes a richer task representation in the form of a goal distribution. [15] learns a goal conditioned policy by maximizing the mutual information between the goal state and the achieved state. While the goal policy achieves a good performance, the heuristic used for goal sampling is very simple and was not shown to scale to large environments with distractors. [6] improves upon [5] by designing a mechanism that incentivizes the goal sampling mechanism to focus on the frontier of the distribution of known goals, implementing a form of novelty search [29]. While these approaches efficiently push exploration from the initial location, they can be naturally attracted by distractors that generate unpredictable or uncontrollable novel observations (either due to randomness, lack of cognitive capacity or lack of access to hidden information), and in general do not construct a curriculum based on and optimized for the agent's current performance. Here, we experimentally identify this limit, and show that the learning progress based goal sampling mechanism we introduce can be used as a wrapper around these algorithms to enable them to avoid irrelevant and unlearnable regions of the environment, while directing their efficient novelty-based exploration to the interesting regions.

**Curiosity-driven Deep RL** A common trend to improve exploration in the classical reinforcement learning setting with sparse external rewards has been to supplement the task reward with intrinsic reward [30]–[32]. Approaches considering image-based low-level perception have used intrinsic rewards measuring various forms of novelty, based on counts [30] or prediction errors [31], [33]. While these approaches can give impressive results, they have fallen short on simple environments in the presence of a distractor that is partially controlled by one of the agent's actions [34]. [35] studied the use of learning progress as an intrinsic reward to enable Deep RL agents to be robust to distractors. However, this approach relied on high-level disentangled state representations and did not include the notion of goal exploration.

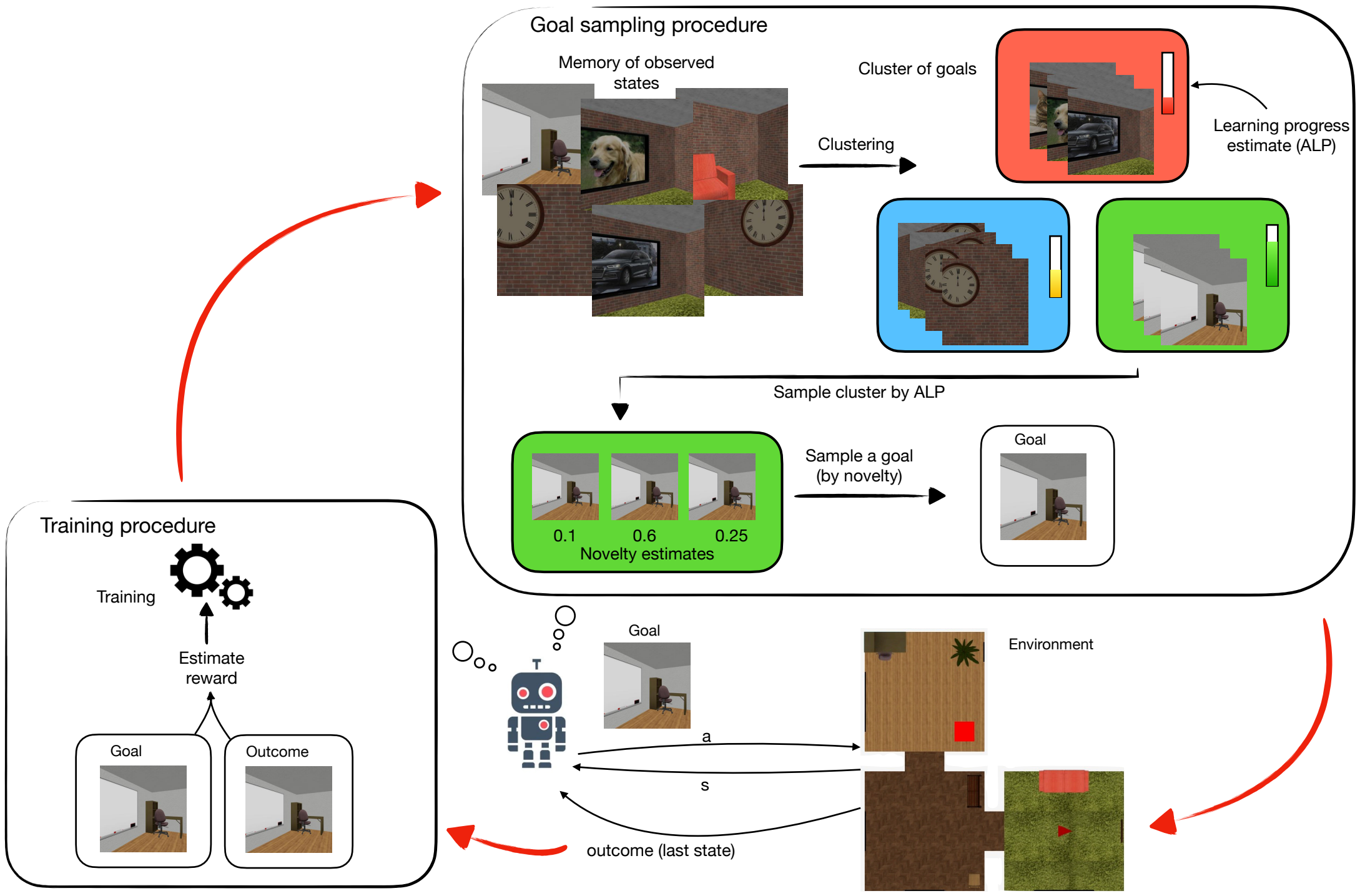

Fig. 1: This figure depicts the GRIMGEP approach to goal exploration. The agent clusters all the observed images and estimates ALP for each cluster. Based on this estimates, it then samples a cluster. Then, it estimates the novelty of all the goals inside this cluster, and samples a goal using these estimates. The agent then interacts with the environment and attempts to reach the goal. When the episode is over, the agent is trained using the reward it itself estimated. During the whole process, no expert knowledge is used.

A parallel line of work on curriculum learning is that of gradually constructing training environments suited to the agent's competence. In ADR [36] an agent is trained on a parameterized environment. Using expert knowledge, the parameter set is graduarly expanded at the boundaries when the agent reaches a sufficient performance. In PAIRED [37] three agents are trained: the adversary, the protagonist, and the antagonist. The adversary is trained to construct environments maximizing the antagonist's reward (to ensure feasibility) while minimizing the protagonist's reward (to make the current policy more robust). Another example is POET [38], [39], which jointly evolves a population of pairs (environment, agent).

Learning progress estimates have also been used for other purposes. CD-MISFA [40] uses ALP choose the optimal context to improve representation learning of classical RL agent. IM2C [41] uses LP estimates to choose between a model-based and a model-free controller, and use negative LP as an intrinsic reward to train the DRL agent. GRAIL [42] uses LP estimates to choose an image goal. Compared to this paper, they use a different setting: linear agents that receive joint locations as inputs, and do not consider noisy distractors.

## III. Problem Definition

### A. Open-ended unsupervised exploration in image based environments

In open-ended unsupervised exploration, the agent has to acquire a diversity of skills without access to the internal dynamics of the environment (expert knowledge). The only information the agent receives from the environment are the observations. The agent must autonomously discover new states and learn how to reach them. In practice, the agent often learns a goal-conditioned policy, and has an internal goal sampling mechanism (an autotelic agent).

The lack of expert knowledge makes the problem challenging in two ways. First, the agent has to autonomously train the goal-conditioned reward function (no measure of similarity between a state image and the goal image is given). This step is challenging as the distance of two images in the pixel space (ex. L2) doesn't correspond to their semantic similarity. Second, a mechanism must be constructed to autonomously create and sample plausible goals (no information is given about which images represent feasible states in the environment).

### B. Intrinsically motivated goal exploration

Intrinsically motivated goal exploration processes (IMGEP) are a natural way to address this problem. IMGEPs are a family of approaches that explore by repeatedly setting and trying to achieve goals.

Those approaches usually consist of two parts: a goal reaching policy, and a goal sampling mechanism. A good policy ensures that goals are reached efficiently, and the goal sampling mechanism discovers new goals. The combination of these two elements ensures that the agent explores the environment efficiently. In this work, we focus on the goal sampling mechanism.

In image-based environments with no expert knowledge, goals cannot be sampled directly from the image space, as such images are unlikely to represent feasible goals.

A simple solution is to sample goals from the history of observations. However, this history will often be disproportionately filled with observations that are close to the starting position of the agent. Or more generally, by observations that correspond to states which are easy to reach. Thus, sampling goals uniformly from the history of observations is not an efficient strategy to explore. To find interesting goals, current state-of-the-art methods estimate the novelty for all possible goals in order to maximize the diversity of sampled goals. A pseudocode of such an algorithm is shown in Algorithm 1 in black lines.

A drawback of novelty-seeking approaches is that they sample goals based on novelty alone and don't take into account the feasibility of the goal. In other words, they are not designed to construct a curriculum based on the agent's capabilities. One manifestation of this drawback is in environments with distractors, where unpredictable (possibly structured) images can be perceived (ex. a TV continuously showing novel images). Such distractors are problematic because a randomly occurring goal is both infeasible and novel. Hence, novelty seeking approaches can be attracted to infeasible goals, resulting in reduced performance and, often, catastrophic forgetting.

IMGEPs using Absolute Learning Progress (ALP) have been shown to address both of those issues in simpler (not image-based) environments. This paper, introduces the **GRIMGEP** approach that extends ALP based approaches to unsupervised image-based exploration, and combines them with novelty-based approaches. GRIMGEP uses ALP estimates to detect learnable regions of the goal space, and then uses novelty-based approaches to sample goals inside these regions.

### C. Evaluation

The objective evaluates the performance of the goal-reaching policy over a test set of goals, which represent the diversity of learnable skills in the chosen environment (the learner has no access to this test set):

$$\text{argmax}_{\theta} \int_{g \sim \mathcal{G}} f(g, \tau_{\pi_\theta, g}) \, \mathrm{d}g. \quad (1)$$

Where $\mathcal{G}$ is the goal space (in practice the test set), $\tau_{\pi_\theta, g}$ the trajectory resulting from following the goal-conditioned policy $\pi$ (parameterized by $\theta$) aiming for goal $g$. $f$ is the goal fulfillment evaluation function specifying to what extent was the goal $g$ reached in episode $\tau_{\pi_\theta, g}$. This function is defined by the experimenter for the purpose of evaluation, and the agent has no access to it during training.

In our case $f$ is a measure of similarity between the goal and the final state (i.e. to what extent are the same objects visible on the two images). Since it is unfeasible to integrate over the whole image space, a fixed set of testing goals (test set) is used for evaluation. For further details on the evaluation and the evaluation metric, refer to section VI-A.

**Algorithm 1:** High level pseudocode of the Intrinsically Motivated Goal Exploration Process (blue lines correspond to GRIMGEP additions)

```
// the goal-conditioned policy and the
   goal conditioned policy's reward
   function (VAE)
π.init() R.init()
encountered_states = []
performance_history = []
// random rollouts to fill the replay
   buffer
for N_warmup do
    traj = env.random_rollout()
    encountered_states.add_states(traj)
// exploration
for ep in N_epochs do
    if ep >start_exploration then
        clusters = clustering_fn(performance_history)
        per_cluster_alp = estimate_ALPs(clusters,
          performance_history, R)
        // sample a cluster and construct
           the prior
        prior=construct_prior(clusters, per_cluster_alp)
        // novelty-based
        goal_intrinsic_rewards = compute_intrinsic_rewards(
          encountered_states)
        goal_intrinsic_rewards *= prior      // by eq. 5
    else
        // at the beginning sample uniformly
        goal_intrinsic_rewards = [1,1...1]
    goal = prioritized_sample_goal(encountered_states,
      goal_intrinsic_rewards)
    trajectory = env.policy_rollout(π, goal)
    encountered_states.add_states(trajectory)
    performance_history.add_pair((goal, trajectory.last_state))
    R.train() π.train()        // fit VAE and policy
    // fit PCA and GMM
    clustering_fn.train(encountered_states)
```

## IV. Current novelty-based exploration approaches

We consider two representative novelty seeking goal exploration approaches: Skewfit [6] and CountBased. Both Skewfit and CountBased train a generative model ($\beta$-VAE) online on the encountered states. They both use the negative $L2$ distance between the goal and the state in this latent space as the reward for training the goal-conditioned policy. They both train a goal-conditioned policy using SAC [13] with HER [12]. The reward used is the current VAE's negative L2 latent distance between the goal and the state. The architecture of the policy is taken from [6]. Both Skewfit and CountBased sample goals from a set of all the encountered states, however they prioritize goal sampling differently.

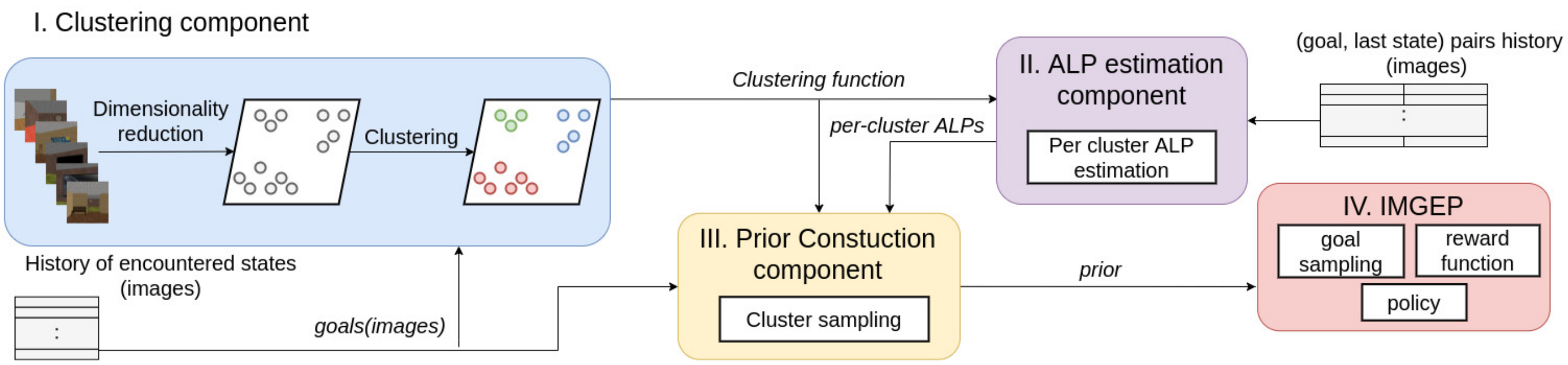

Fig. 2: Goal sampling procedure in the GRIMGEP framework. 1) The **Clustering component** clusters the goal space into different components. In practice, the possible goals are the encountered states, and so clustering is performed on the *history of encountered states*. 2) **ALP Estimation component** computes the learning progress of each cluster using the *"(goal, last state)" pairs history* (the history of all attempted goals and their corresponding outcomes). 3) **Prior Construction component** samples a cluster using the ALP estimates, and constructs the goal sampling prior as the masking distribution. The prior distribution assigns uniform probability to goals from the *history of encountered states* that belong to the sampled cluster, and 0 probability to goals outside the cluster. 4) The **Underlying IMGEP** samples a goal from the distribution formed by combining the goal prior and the underlying IMGEP's novelty-based goal sampling distribution: a novel looking goal is sampled from the sampled cluster. Further details, are given in Section V.

Skewfit [6] outperformed many baselines for unsupervised goal exploration in environments without noisy distractors. Goal sampling is prioritized based on how novel they appear to the current generative model (VAE): the probability of sampling a goal is inversely proportional to the probability of that goal under the generative model's distribution. For a more detailed description of Skewfit we refer the reader to section A-A in the supplementary. As Skewfit tends to behave unstably when sampling purely based on the novelty bonuses, the authors introduced a regularization hyperparameter $\alpha$. $\alpha$ enables interpolation between a uniform ($\alpha = 0$) and a completely exploratory distribution ($\alpha = -1$) making Skewfit more robust. We consider two versions of Skewfit corresponding to different values of $\alpha$: 0.25 and 0.75.

CountBased's goal sampling is inspired by how Go-Explore samples cells [11]. Observations are downsampled to a 3x3 image, and quantized to 4 values ($4^3$ different colors). The number of times an observation has been encountered is counted as the number of time the downsampled version of the observation has been encountered. The probability of sampling a goal is then defined to be inversely proportional to the number of times its quantized version was encountered. For a more detailed description of CountBased we refer the reader to section A-C in the supplementary.

## V. GRIMGEP

We present the **G**oal **R**egions guided **I**ntrinsically **M**otivated **G**oal **E**xploration **P**rocess (**GRIMGEP**) framework. The GRIMGEP framework extends any goal exploration algorithm with an ALP based curriculum. By doing so, it combines the robustness in the absence of distractors of SOTA novelty-based algorithms in the image space with the more powerful curricula of the ALP-based approaches. It improves the goal exploration algorithms by identifying interesting and learnable regions of the goal space and by sampling goals from those regions.

Clustering has been previously used to improve ALP estimation [28], [43], but not on the image space. To do this, we use the feature extractor of a pretrained object detector to encode image goals to a low dimensional latent space. The idea is that clusters corresponding to regions that are either unlearnable, already learned, or too hard for the agent capabilities will have a low ALP and will thus be sampled rarely. On the contrary, regions that are currently being learned or forgotten (performance rising or falling) will have high ALP and will be sampled more. The agent will thus autonomously build a curriculum of tasks, gradually moving from easier tasks to harder tasks.

In a nutshell, GRIMGEP extends the ALP approach to image-based exploration by: 1) fitting a clustering mechanism (PCA and GMM) on the embeddings of all encountered images (embeddings are created with a pretrained feature extractor), 2) clustering the history of (attempted goal, achieved last state) pairs, 3) estimating per-cluster ALP, 4) and sampling a cluster based on these ALP estimates. The wrapped IMGEP is then used to sample a goal from the history of all encountered states, but constrained to goals that belong to the sampled cluster. This pipeline is repeated for every sampled goal. The extensions introduced by GRIMGEP over classical novelty based IMGEPs are shown at a high level by blue lines Algorithm 1.

### A. GRIMGEP components

The GRIMGEP framework consists of the following four components, as shown in Figure 2:

**I. Clustering component.** It creates a clustering function that maps any goal to a cluster. The clustering function is depicted in Figure 3. It consists of two parts: a *dimensionality reduction*, and a *clustering mechanism*.

*The dimensionality reduction* starts with the feature extractor (backbone) taken from a pretrained object detector (YOLO-V3 [44]), and is followed by global average pooling [45], this feature extractor is frozen and is never trained. We use a pretrained object detector because it provides a latent space suitable for clustering different realistic objects, however this constrains our approach to realistic 3D environments. YOLO-v3, which we obtained from the GluonCV library [46] ('yolo3_darknet53_coco'), was pretrained on the COCO [47] object detection dataset. The dimensionality reduction reduces a $48 \times 48 \times 3$ image to a $1 \times 1 \times 1024$ vector.

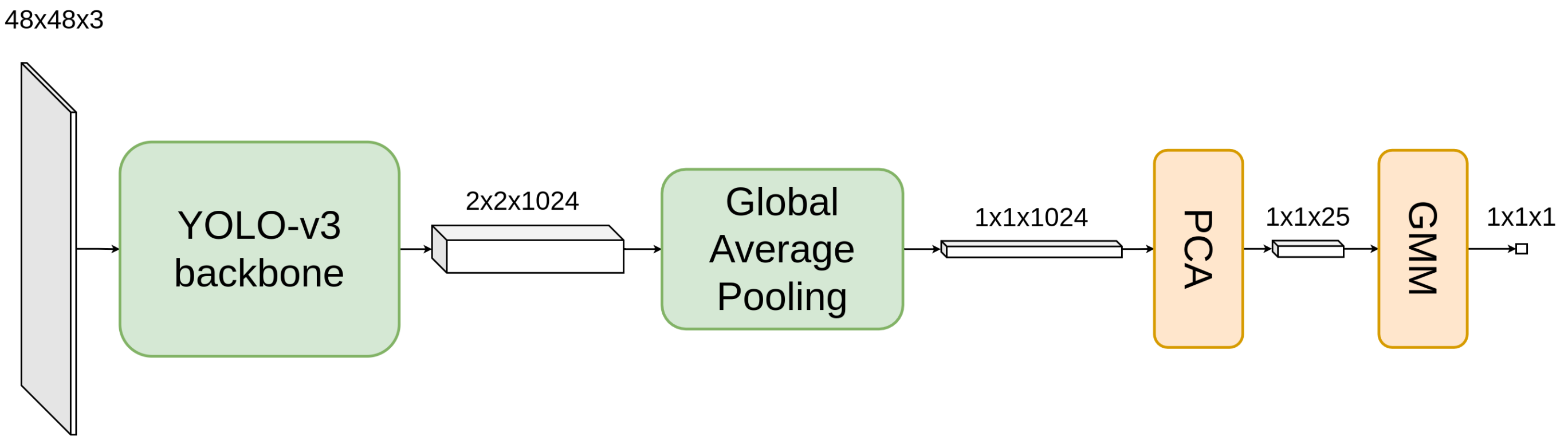

Fig. 3: The clustering function: it consists of a dimensionality reduction, and a clustering mechanism. The pretrained dimensionality reduction (green) is not trained and produces encodings. The clustering mechanism (orange) is fit from scratch every epoch using the encodings of all the encountered images.

*The clustering mechanism* consists of a PCA [48] and a GMM [49]. Those two components are fit every epoch on the encodings of all the encountered images.

This component produces a *clustering function* that maps any image to a cluster_id. This function is constructed anew every epoch and passed to the *ALP estimation component* and the *Prior construction component*. Those components will use it to estimate ALP and sample a suitable cluster. Many other works used clustering to estimate ALP efficiently in low dimensional spaces: in this work we study how the clustering approach can be transposed in the image space.

**II. ALP Estimation component.** This component is used to estimate the absolute learning progress for each cluster.

To estimate learning progress, we need a history of performances. The ALP Estimation component keeps a history of attempted goals, and the last achieved states (pairs of images). Only the most recent $l$ entries are used. The performance for each pair is computed using the reward function of the underlying IMGEP (negative L2 distance in the VAE latent space).

The clustering function assigns a cluster_id to every sampled goal. The ALP is then estimated for each cluster using the following equation:

$$ALP_c = \left|\frac{1}{l/2}\sum_{i=L-l}^{L-l/2} h_{i,c} - \frac{1}{l/2}\sum_{i=L-l/2}^{L} h_{i,c}\right|, \quad (2)$$

where $L$ is the full cluster's history length, $l$ is the maximum considered length, $h_i$ is the cluster's performance history at epoch $i$ (the average performance for goals sampled from cluster $c$ in epoch $i$).

Pairs of (goal, last_state) are grouped based on the cluster to which the goal belongs to. Since the reward function is the $L_2$ distance in the latent space of a VAE that is trained during exploration, we use the current state of the VAE to compute the performance. To construct the per-cluster performance history, we average the performances corresponding to goals sampled

**Algorithm 2:** ALP estimation component

```
def estimate_ALP(cluster_ids, perforance_history, R) :
    // Initialize an empty history for
       every cluster
    for c in N_clusters do
        h[c] = []

    // Estimate pefromances epoch by epoch
    for epoch ∈ epochs_so_far do
        // Initialize an empty performance
           history for every cluster
        for c in N_clusters do
            p[c] = []

        // Extract the (goal, last_state)
           pairs collected in the epoch
        epoch_rollouts = performance_history[epoch]

        // compute the performances from the
           epoch for each cluster
        for rolloutinepoch_rollouts do
            goal_c = cluster_ids[rollout.goal]
            // recompute the rewards
            goal_perf = R(rollout.goal, rollout.last_state)
            p[goal_c].append(goal_perf)

        // add the mean performances from
           this epoch to the cluster's
           history
        for cinN_clusters do
            h[c].append(mean(p[c]))

    // LP estimation
    for c in N_clusters do
        ALP_c = estimate_ALP(h[c]) ; // using Eq 2
```

in the *same epoch* and assigned to the *same cluster*. Only the estimates from the last $l$ epochs are kept. These histories are then used to estimate absolute learning progress for each cluster by Eq. 2. This component is further explained in Algorithm 2.

This component produces an ALP estimate for each cluster. These estimates are then used in the *Prior construction component* to sample a cluster.

**III. Prior Construction component.** This component sam-

ples a cluster and constrains the Underlying IMGEPs' novelty-based sampling to this cluster. We do this by constructing a prior distribution and passing it to the Underlying IMGEP.

The prior distribution is constructed in two steps. First, a cluster is sampled proportionally to the per cluster ALP estimates. Then, the prior distribution is defined as a masking distribution: uniform over all goals that the clustering function would assign to the sampled cluster, and zero elsewhere.

More formally those two steps are as follows:

1. Using per-cluster ALP estimates, sample a cluster according to the probability defined in the following equation:

$$p(c) = \frac{4}{5}\frac{LP_c^T}{\sum_{i=1}^{C} LP_c^T} + \frac{1}{5}\frac{1}{C}, \tag{3}$$

   where $C$ is the number of clusters and $T$ is a hyperparameter. The hyperparmaeter $T$ enables us to set how much priority we want to give to the high ALP clusters, which is usefull if we have a lot of clusters with small LPs.
2. Construct the prior according to the following equation:

$$prior(g) = \begin{cases} \frac{1}{n_c}, & cl(g) == c \\ 0, & \text{else} \end{cases}, \tag{4}$$

   where $c$ is the sampled cluster, $n_c$ is the number of goals from the replay buffer that would be assigned to cluster $c$, and $cl$ is the clustering function assigning any state to a cluster. This is essentially a masking distribution giving uniform probabilities for goals inside the sampled cluster and zero for the rest.

The Underlying IMGEP multiplies its sampling distribution with the prior distribution. This constrains the goal sampling to the sampled cluster.

**IV. Underlying IMGEP.** This component can be any goal-exploration process that constructs a goal sampling distribution. Goals are sampled and given to the agent. The agent is trained on reaching those goals and the process is repeated.

When used outside the GRIMGEP framework, goal sampling is performed according to a novelty maximizing distribution $p_{imgep}$.

When used inside the GRIMGEP framework, the goal sampling distribution $p_{imgep}$ is combined with the prior distribution to define the final goal sampling distribution $p$ as follows:

$$p(g) = \frac{p_{prior}(g)p_{imgep}(g)}{\sum_{g \in R} p_{prior}(g)p_{imgep}(g)}, \tag{5}$$

where $R$ is the set of all encountered states.

Since $p_{prior}$ is zero everywhere outside the sampled cluster, and $p_{imgep}$ prioritizes novel goals, the resulting distribution $p$ will select a novel goal inside the sampled cluster.

In this work, we study two underlying IMGEPs: Skewfit and CountBased. Both sample novel looking previously observed states (images) as future goals. Both train a RL policy using the negative $L2$ distance of an online trained VAE as rewards.

**Algorithm 3:** Implementation details of the Intrinsically Motivated Goal Exploration Process in the Deep RL setting (blue lines correspond to GRIMGEP additions)

```
π.init() R.init()  // goal-conditioned policy and
  the goal conditioned policy's reward
  function (VAE)
encountered_states = []
performance_history = []          // A history of
  (sampled_goal, last_state) pairs
// random acting rollouts to fill the
   replay buffer
for N_warmup do
    traj = env.random_rollout()
    encountered_states.add_states(traj)
// exploration
for ep in N_epochs do
    for N_parallel_processes do
        if ep >start_exploration then
            // novelty-based exploration of
               the underlying IMGEP
            goal_intrinsic_rewards =
             compute_intrinsic_rewards(encountered_states)
            // Assign a cluster_id to every
               goal from the
               perfromance_history
            cluster_ids = clustering_fn(performance_history)
            // Recompute the performances and
               estimate per-cluster ALP
            per_cluster_alp = estimate_ALPs(cluster_ids,
             performance_history, R)     // by Alg. 2
            sampled_cluster=sample(clusters, per_cluster_alp)
             // by Eq. 3
            // Create a masking distribution
               by Eq. 4
            prior = construct_prior(sampled_cluster,
             encountered_states)
            // Integrate the bias by Eq. 5
            goal_intrinsic_rewards *= prior
        else
            goal_intrinsic_rewards = [1,1...1]  // at the
             beginning sample uniformly
        goal = prioritized_sample_goal(encountered_states,
         goal_intrinsic_rewards)
        trajectory = env.policy_rollout(π, goal)
        encountered_states.add_states(trajectory)
        performance_history.add_pair((goal,
         trajectory.last_state))
    R.train() π.train()      // fit reward function
     (VAE) and policy
    clustering_fn.train(encountered_states)    // fit PCA
     and GMM
```

### B. Implementational details on the GRIMGEP approach

A detailed pseudocode is shown in the Algorithm 3. First, we do a *warm up* phase to fill the replay buffer by doing random actions. Then, for the first $start_exploration$ epochs, we do not use intrinsic rewards, we just sample goals uniformly from the replay buffer. After this, we start using intrinsic rewards to prioritize the goal sampling. We use parallel computing by sampling ten goals per epoch. For each of those goals, we sample a new cluster, construct a new prior and then sample a new goal using the probability distribution obtained by combining the prior and the goal sampling probability distribution of the IMGEP. This procedure is detailed in Algorithm 3, where the blue colors depict GRIMGEP addition.

The biased distribution, $p$ which was constructed by Prior construction component, is used everywhere where the underlying IMGEP would normally use $p_{imgep}$. Therefore, for three purposes: 1) sampling goals for exploration, 2) sampling replacement goals for HER, and 3) since all of our underlying IMGEPs train their own VAE, biasing the training of this VAE.

## VI. Experiments

In this section, we present experiments studying the effect of the GRIMGEP framework[2] on two novelty-seeking goal exploration processes: Skewfit [6] and CountBased (inspired by GO-Explore [11]). In order to assess the effect of the GRIMGEP framework, we compare the behavior of those algorithms when used alone and when combined with the GRIMGEP framework. Experiments are performed in a 3D first-person environment containing a distractor.

As discussed earlier, noisy distractors are just one of many problems that can be addressed by constructing an efficient curriculum. For instance, an efficient curriculum also allows the agent to avoid already learned or currently too hard tasks. It enables the agent to gradually build on the knowledge acquired on easier tasks to solve harder tasks. In simpler scenarios, ALP has been shown to help with many aspects of exploration [9], [28].

While evaluating the impact of learning progress in simple scenarios can be safely assessed by measuring the sampling rate on easy, medium, hard, and impossible tasks, precisely evaluating and measuring exploration in a rich 3D environment with many objects is a difficult task. In the supplementary, see section VI-C for a discussion, and section D for experiments in a toy environment. In order to study how learning progress can be used in rich 3D environments, we focus on the noisy TV problem as it encapsulates most of the features shared by problems solved by the learning progress (identifying different tasks, ignoring tasks that are not relevant and focusing on interesting ones).

More precisely, we study the following questions:

- How do current approaches behave in the presence of action-induced distractors?
- How does the GRIMGEP framework change the behavior of current approaches?
- How important are the ALP estimates for the performance of the GRIMGEP framework?

[2]We used and adapted the following codebases in our experiments: https://github.com/vitchyr/rlkit, GluonCV [46], sklearn [50]

### A. Explore3D environment

Explore3D is a 3D first-person environment implemented using the MiniWorld [51] environment simulator. The environment consists of 14 different objects located in three separate rooms (TV room, Gallery room, Office room). The topology of the environment and the different rooms are depicted in Fig. 4. The agent's observation space consists of RGB images depicting the first-person view of the environment, as shown in Figures 4b-4d. The agent can interact with the environment through 7 continuous actions: four for movement, two for rotating the camera and one for turning the TV on or off. The starting position is in the TV room facing the clock, as depicted by the red triangle in Fig. 2a.

At the beginning of each episode, the TV is turned OFF (a black screen). When the agent turns on the TV, a random image from the ImageNet [52] validation set is shown on the screen. While turned on, the TV shows a new image every step with a probability of 0.1.

The environment provides a challenge for exploration in the following two aspects: 1) if no explicit exploration incentive is used, the agent will not be able to explore further than the starting TV room, and 2) novelty based exploration will be attracted to the diversity of images on the TV. Furthermore, the reaching goals depicting a particular image on the TV is almost impossible as the agent has no control over which image, from the 50000 ImageNet validation images, will be shown on the screen.

### B. Evaluation

For evaluation, we construct a test set consisting of 20 goal images. Those images depict all the objects in the environment except the activated TV.

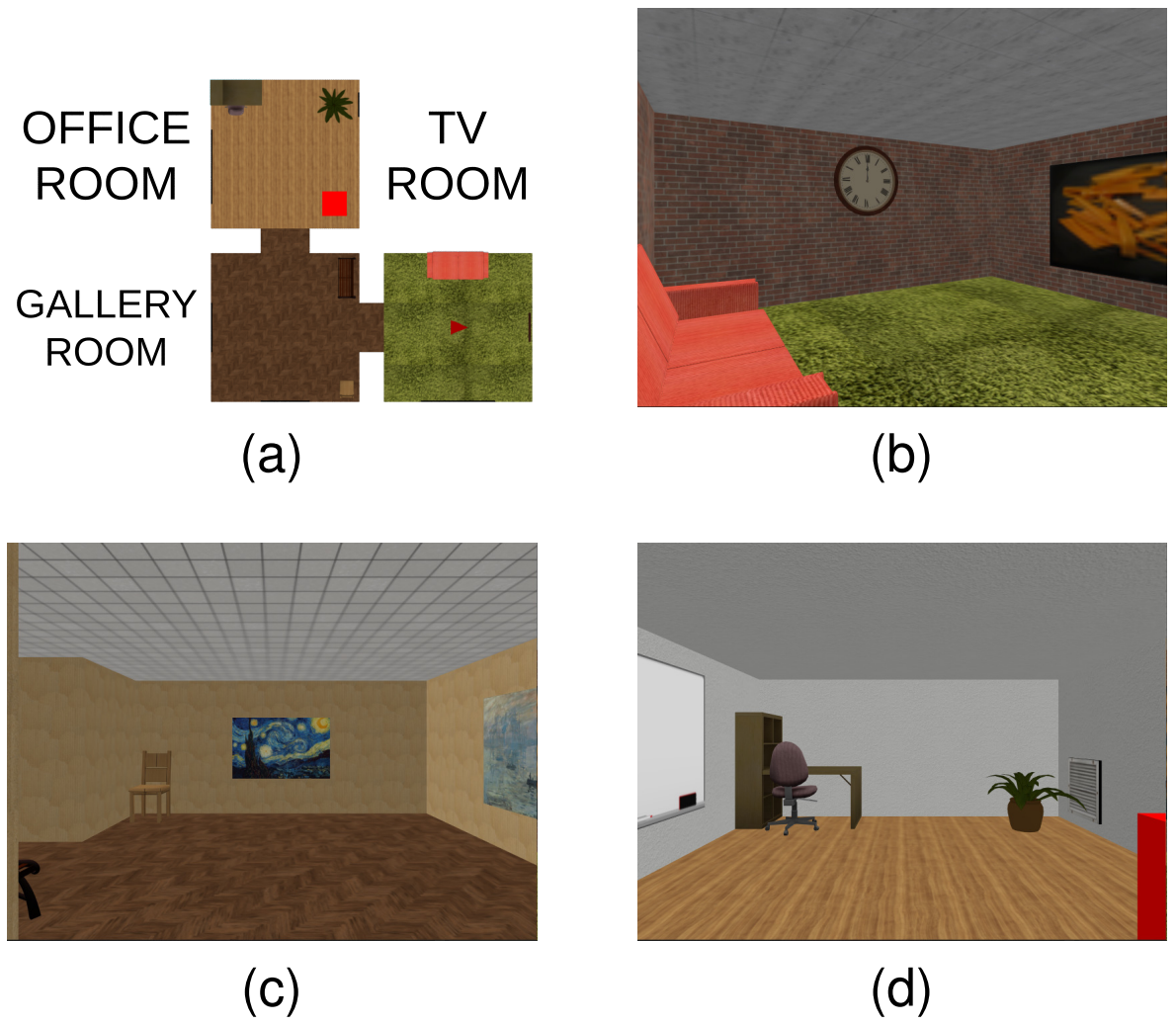

Fig. 4: Different rooms of the 3D environment (a) Topology (b) TV room (c) GALLERY room (d) OFFICE room

To evaluate the fulfillment of a goal, we construct the *Visible entities F1 score* metric. This metric is only used in evaluation. Using the environment simulator, we extract a list of visible objects (entities) for both the goal and the last state from the episode. By treating objects visible in the goal as ground truth, precision and recall, and finally F1 score are calculated.

This metric, although it captures well the task of "seeing the goal image", is not perfect. We can notice that it is insensitive both to the angle and the distance at which the agent is looking at the objects. Furthermore, an object may be visible on the state image with only a pixel or two. This object will still be treated as fully visible, even though it's very hard for the agent to compensate for this error. Nonetheless, we found the metric sufficiently accurate for comparing different algorithms.

### C. Tracking exploration in rich 3D environments

In simple environments where different tasks are predefined, previous works has demonstrated quite precisely how ALP estimates guide exploration from easier tasks to harder tasks (which often build on easier tasks), while at the same time avoiding unlearnable tasks and repeating previously learned tasks which have started to be forgotten. For example, in [9] an agent first learns to grab blocks, and then to stack them.

When one considers rich environments with a variety of objects, as we do, this precise tracking becomes quite challenging. Where before was a clearly defined task of "grabbing blocks", we have a cluster in a PCA reduced latent space of a pretrained object detector.

It is not trivial to interpret even what one such cluster refers to, and much harder to construct a way in which to track sampling of it. Each epoch, new clusters are constructed on a new PCA dimensionality reduction embeddings. Therefore, the semantic meaning of the clusters changes during training in an unpredictable and untrackable way.

There are a few ways in which one might try to loosely estimate exploration, however they are not as useful as they might at first seem. One might try to track sampling goals in different locations (rooms), or sampling of goals depicting different objects. Tracking per-room sampling is sufficient for our preliminary 2D environment (see section D in the supplementary), where different rooms were constructed intentionally to correspond to tasks of different difficulty. However, in the 3D environment, the location of a goal says little of its difficulty. For example, if we consider two goals: one of looking at an object frontally, and another of looking at an object from a specific angle in the same room while also observing a specific object in the background. Both of those goals are in the same room, but they are of significantly different difficulty. Another example is that of two goals looking at the same object, but one from proximity and another through the door from a different room. This time both goals show the same object, but are again of very different difficulty.

### D. Results

In the following section, we present the experiments with Skewfit and CountBased. We refer to the GRIMGEP wrapped versions of those algorithms as GRIM-Skewfit and GRIM-CountBased.

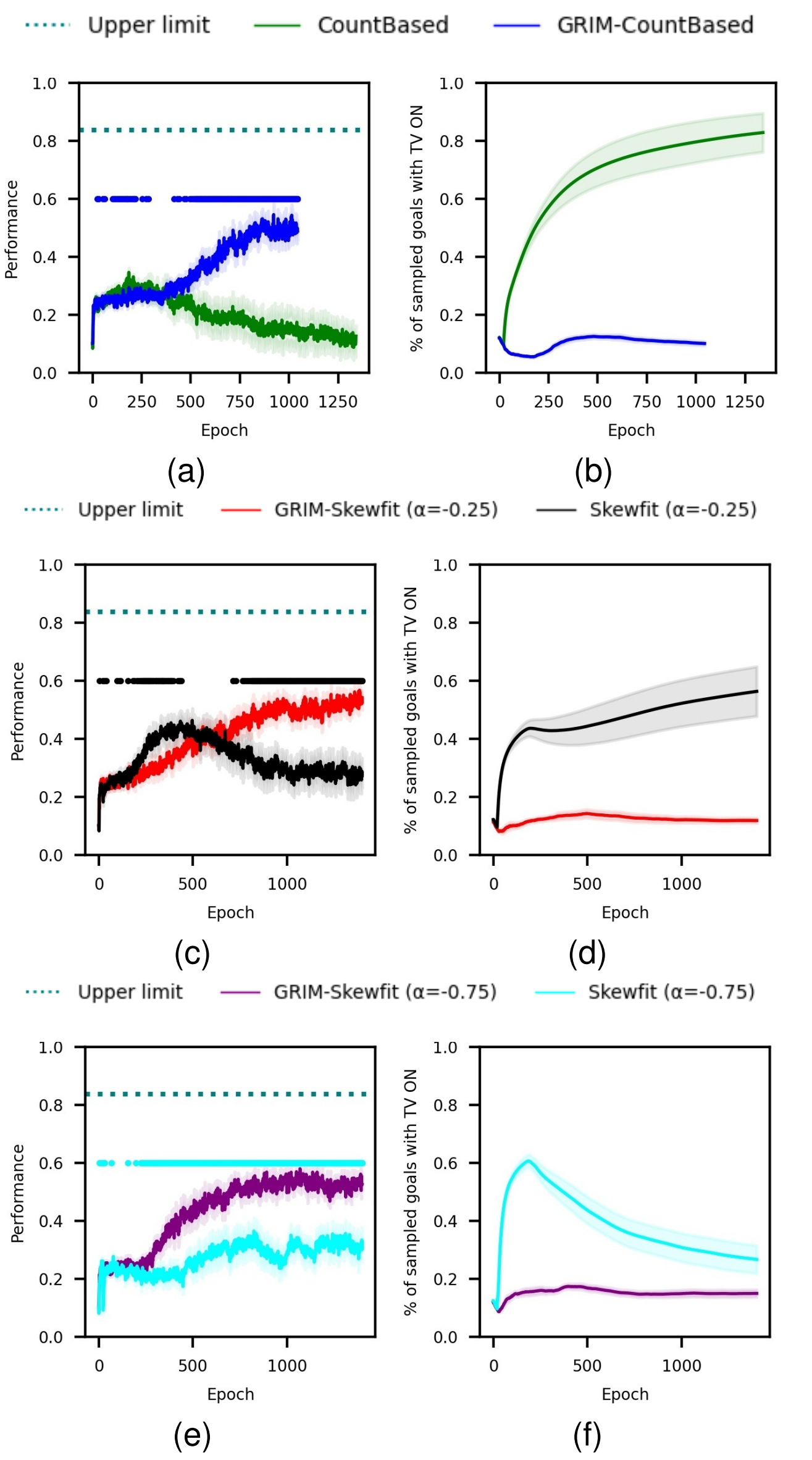

Fig. 5: Performance (left) and percentage of all sampled goals depicting the active TV($\pm$ std. err.) (right), shown for CountBased and Skewfit with two different values of $\alpha$, compared to the three approaches wrapped inside the GRIMGEP framework. We can see that the performance of all the three approaches is improved, and catastrophic forgetting avoided, when wrapped inside the GRIMGEP framework. The dots depict statistical significance (Welch t-test, p=0.05, 15 seeds).

**How do current novelty-based approaches behave in the presence of noisy distractors?**

Figure 5 shows the performance of CountBased and Skewfit with two different values of hyperparameter $\alpha$, along with the corresponding percentage of all sampled goals depicting the active TV.

The horizontal line shows the best performance achieved by any of these approaches (CountBased) in a setting where the TV couldn't be turned on. This line represents an upper bound to the performance. As we can see, all the baselines oversample goals depicting the active TV, thereby significantly limiting their performances. Furthermore, the performance of CountBased and Skewfit ($\alpha = 0.25$) diminishes as they continue to oversample those goals, resulting in *catastrophic forgetting*.

These experiments show that the presence of a noisy distractor is challenging for the novelty-based exploration of both Skewfit and CountBased. This motivates the use of more advanced exploration techniques such as GRIMGEP.

**How does the GRIMGEP framework change the behavior of current approaches in the presence of noisy distractors?**

Figure 5 shows the performance and the corresponding percentage of all sampled that depict the active TV of CountBased and Skewfit with two different values of hyperparameter $\alpha$ when wrapped inside the GRIMGEP framework (prefixed by "GRIM-").

In figure 5 we can see that all approaches achieve greater performances compared to their unwrapped versions. Furthermore, we can see that GRIM-CountBased and GRIM-Skewfit ($\alpha = 0.25$) have overcome the problem of *catastrophic forgetting* present in their unwrapped versions. We argue that this is because both GRIM-Skewfit and GRIM-CountBased sample the TV considerably less than their unwrapped versions.

These experiments show that wrapping the novelty seeking approaches with the GRIMGEP framework alleviates some problems introduced by noisy distractors, constructs a much better curriculum, and achieves a much higher overall performance.

**How important are the ALP estimates for the performance of the GRIMGEP framework?**

We do an ablation study on the cluster sampling mechanism (prior construction component). In figure 6, we compare two different ways of sampling clusters: according to their ALP estimates (prefix: "GRIM-ALP"), and uniformly (prefix: "GRIM-UNI").

In Figures 6a and 6b we can see that even though the clustering by itself is enough to limit the oversampling of the TV, ALP is still key for the final performance. For CountBased, ALP is needed to construct a curriculum and move the exploration far away from the starting position.

In Figures 6c, 6d, 6e and 6f we can see that for Skewfit the ALP estimates do not add much to the overall performance. The clustering mechanism is not only sufficient to escape the distractor, but also to enable Skewfit to create an adequate curriculum.

In conclusion, while the GRIMGEP's clustering mechanism alone successfully limits the sampling of the distractor, this is only the first step in constructing an adequate curriculum. The scale of the benefits provided by ALP estimates depend on the efficiency of the underlying exploration algorithms. This observation is consistent with previous observations made in state based environments, which showed that separating the goal spaces into conceptually different parts and sampling goals from those regions uniformly is in general already a strong baseline that outperforms naive sampling in the goal space [9], [14], [19].

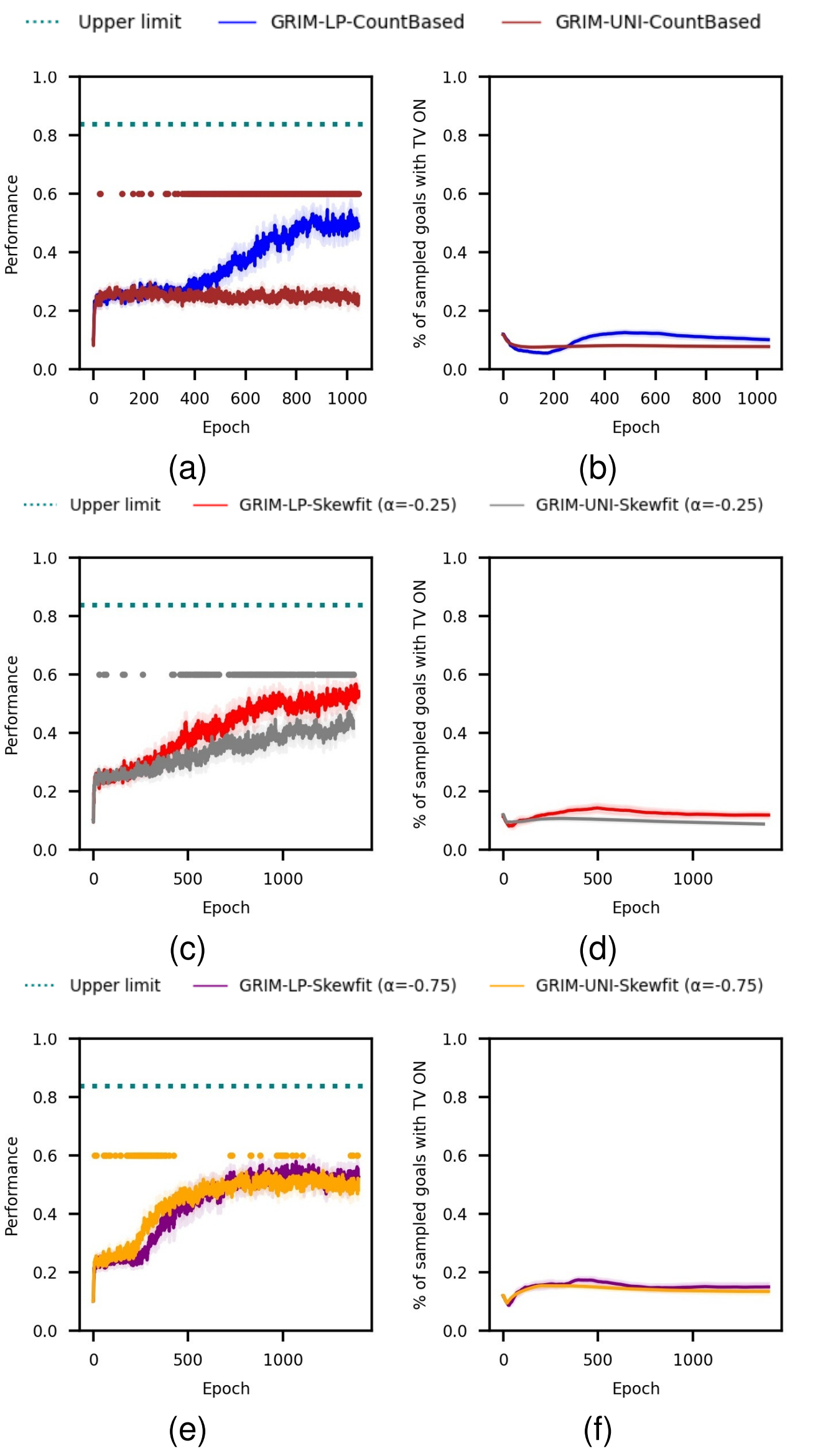

Fig. 6: The performance and the percentage of sampled goals depicting the active TV ($\pm$ std. err.) The dots depict statistical significance (Welch t-test, $p = 0.05$, 15 seeds) (a) Performance CountBased (b) % of TV goals CountBased (c) Performance Skewfit ($\alpha$ = 0.25) (d) % of TV goals Skewfit ($\alpha$ = 0.25) (e) Performance Skewfit ($\alpha$ = 0.75) (f) % of TV goals Skewfit ($\alpha$ = 0.75)

## VII. Conclusion

We studied the problem of unsupervised image-based goal exploration in first-person 3D environments with action-induced noisy distractors. Autotelic agents are a common approach for this problem. Those agents that repeatedly set goals, which they then try to reach. These agents often prioritize goal sampling according to some intrinsic reward, such as novelty or absolute learning progress (ALP) estimates.

We argue that a good way to construct an autotelic agent for rich image-based environments is the following: 1) cluster the goal space, 2) estimate absolute learning progress (ALP) for each cluster, 3) sample a cluster according to ALP, 4) from the sampled cluster, sample a goal according to novelty.

This principle (ALP estimation with clustering) has previously been applied many times, but always in scenarios which did not involve image-level perception [9], [19], [28], [53], or which used population based agents in a 2D environment [8]. Often, expert knowledge was given to the agent in the form of providing high-level encoding of the goal space (e.g. coordinates of objects). In this work, we extend the clustering principle to a rich 3D image-based environment, and we use an unsupervised DRL agent with no access to expert knowledge.

Our experiments show that wrapping standard novelty based exploration mechanisms with the GRIMGEP framework - first sampling a cluster with ALP, and then sampling the goal from the sampled cluster using novelty - creates a better curriculum, ultimately resulting in robustness and drastic performance improvements.

This study has also some limitations. Due to the complexity of the environment, there is a limit to how precisely we could track exploration. In simpler environments, one would separately track sampling on easy, medium, hard and impossible tasks. However, in a 3D image-based environment, it is not feasible to assign a difficulty level to each task due to the complexity of the goal space. For example, looking at an object located far from the starting position can be easier than looking at an object located close to the starting position from a specific angle. Furthermore, when using deep learning agents, this difficulty can also be agent specific. For this reason, we only track sampling of the Noisy TV, which we know to be an impossible task.

Furthermore, we only do experiments on one rich 3D environment (in addition to a preliminary 2D environment in the appendix). It would be interesting to test GRIMGEP on a variety of 3D goal exploration environments with distractors, and also on a variety of distractors.

This work opens up various interesting avenues for future work. The framework is open to various modifications: it would, for example, be interesting to test GRIMGEP on a language goal space as in [54], or in text world environments [55]. Here, instead of an image detector backbone, a large language model would be used to provide embeddings for clustering.

Another interesting avenue would be to connect goal exploration approaches to approaches that use intrinsic rewards as exploration bonuses in sparse reward environments (ex. RND [33], RIDE [32], #Exploration [56]). We could use these intrinsic rewards (which were originally directly optimized by SGD) to prioritize goal sampling inside the cluster sampled by ALP (as an alternative to Skewfit's or CountBased's novelty estimates).

## Acknowledgments

This work was partially supported by Ubisoft, Bordeaux, and by ANR grant DeepCuriosity. Experiments presented in this paper were carried out using the PlaFRIM experimental testbed, supported by Inria, CNRS (LABRI and IMB), Université de Bordeaux, Bordeaux INP and Conseil Régional d'Aquitaine (see https://www.plafrim.fr/), and also benefited from access to the Jean Zay supercomputer (grant A0091011996).

# GRIMGEP: Learning Progress for Robust Goal Sampling in Visual Deep Reinforcement Learning - Supplementary

Grgur Kovač, Adrien Laversanne-Finot, Pierre-Yves Oudeyer

## Appendix A
## Details on the underlying IMGEPs

### A. *Skewfit algorithm*

Skew-fit [1] is an iterative algorithm for training a generative model of the *uniform* distribution over the *feasible* part of the goal space (the goal and the state space are assumed to be equivalent). To ensure that all the generated goals are feasible, Skew-fit trains the generative model using only the observed states. Each state is assigned a weight (probability of being sampled for training the generative model). After each epoch, the weight of each state is updated. A state's weight is set to be inversely proportional to the probability of that state under the current generative model ($w_s \propto 1/G(S)$). These, *skewed*, probabilities are used to train the generative model in the next epoch. As the sampling entropy increases in each epoch, the model converges to modeling a uniform distribution.

This algorithm is then applied to unsupervised goal-driven exploration (training a goal-reaching policy) - a VAE is trained using the replay buffer. They propose two mechanisms for sampling goals for the agent: sampling directly from the current VAE ($q_\psi^G$ in [1]), and sampling from the skewed replay buffer ($p_{skewed}$ in [1]). We focus on the latter as we found it worked better (although both mechanisms could be used with the GRIMGEP framework). This is consistent with [1] as they use $p_{skewed}$ for more complex environments.

In every epoch, goals are sampled, episodes run, new data added to the replay buffer, and both the VAE and the policy are trained.

The skewed probabilities are used for sampling goals, training the VAE, and sampling replacement goals in HER. Uniform sampling is used for training the SAC policy. They denote this algorithm Skewfit + RIG though for simplicity we denote it **Skewfit**.

Therefore, the intrinsic reward used to prioritize goal sampling in Skewfit is shown in the following equation:

$$R_{SKF}(image) = p_{vae}(image)^{\alpha}, \quad (1)$$

where $p_{vae}$ denotes the probability of the image under the current VAE (estimated with importance sampling), and $\alpha$ ($< 0$) is a regularization parameter. This is equivalent to the "skewed" option in [1].

### B. *The $\alpha$ hyperparameter*

To reduce the instability of Skewfit, the authors introduced a regularization hyperparameter $\alpha$ which interpolates between a uniform and a skewed distribution (-1 being completely skewed and 0 being uniform). The original weight is exponentiated by $\alpha$ before normalization. They experiment with 4 different values of this parameter (-0.25, -0.5, -0.75, -1.0).

We found that -0.25 achieves better peak performance for Skewfit, but when wrapped inside the GRIMGEP frames -0.75 works better (see Figure **??**). That demonstrates another benefit of the GRIMGEP framework: using Skewfit inside the GRIMGEP framework enables us to use a more aggressive exploration bonuses ($\alpha = -0.75$) without training becoming unstable.

### C. *CountBased approach*

CountBased is very similar to Skewfit, but it prioritizes goal sampling using a different intrinsic reward. This intrinsic reward is computed using a count-based novelty measure. As shown in eq. 2, we downsize an image (to 3x3) and quantize each channel into 4 different values ($4^3$ different colors), then we count how many times the representation was observed. As in Skewfit, we use this reward to prioritize goal sampling, VAE training, and HER replacement goals. When using this approach as part of the GRIMGEP framework we multiply this distribution with the prior as in Eq. **??**.

$$R_{CB}(image) = count(quantize(downsize(image)))^{\alpha} \quad (2)$$

## Appendix B
## Hyperparametres

For the underlying IMGEPs, all the hyperparameters, including the ones for the training VAE, are the same as in [1] in the "Visual Door" experiments, except the $\alpha$ hyperparameter (see Appendix A-A).

Other hyperparameters are shown in Table I.

| T | 5 |
|---|---|
| episode length | 50 |
| $l$ - cluster history length | 50 |
| $d$ - PCA latent size | 25 |
| $k$ - GMM number of clusters | 25 |

TABLE I: Hyperparmaeters.

## APPENDIX C
## ONLINERIG EXPERIMENTS IN EXPLORE3D

To further understand how GRIMGEP works we study it in combination with an IMGEP that samples goals uniformly. OnlineRIG is equivalent to Skewfit with $\alpha = 0$, which means that the next goal is sampled uniformly from the history of observed states.

Figure 1 shows the performance of OnlineRIG alone ("OnlineRIG"), and wrapped with the GRIMGEP framework ("GRIM-OnlineRIG"). We can see that the two approaches do not differ much in their performance. That is because GRIMGEP *redirects* the underlying IMGEP's exploration. If the underlying IMGEP doesn't explore sufficiently (as OnlineRIG) no progress is observed. By observing the TV sampling, we can see that OnlineRIG samples the distractor less than GRIM-OnlineRIG. This is not surprising. The agent start by looking at the Clock, the couch is also very close. Goals depicting the Clock and the Couch are very easy. OnlineRIG repeatedly resamples such goals, even after the agent has mastered them. That is because they are close to the starting position and the replay buffer if full of them. GRIMGEP, however, clusters various images on the TV into multiple clusters, and since no significant progress is achieved (other than in clock and couch goals) the exploration doesn't move from these goals.

These experiments provide further evidence for our argument: since ALP by itself is not sufficient to direct exploration at the low level, a powerful solution is to use ALP to select relevant *regions* of the goal space, and then use a novelty based approach inside the selected region.

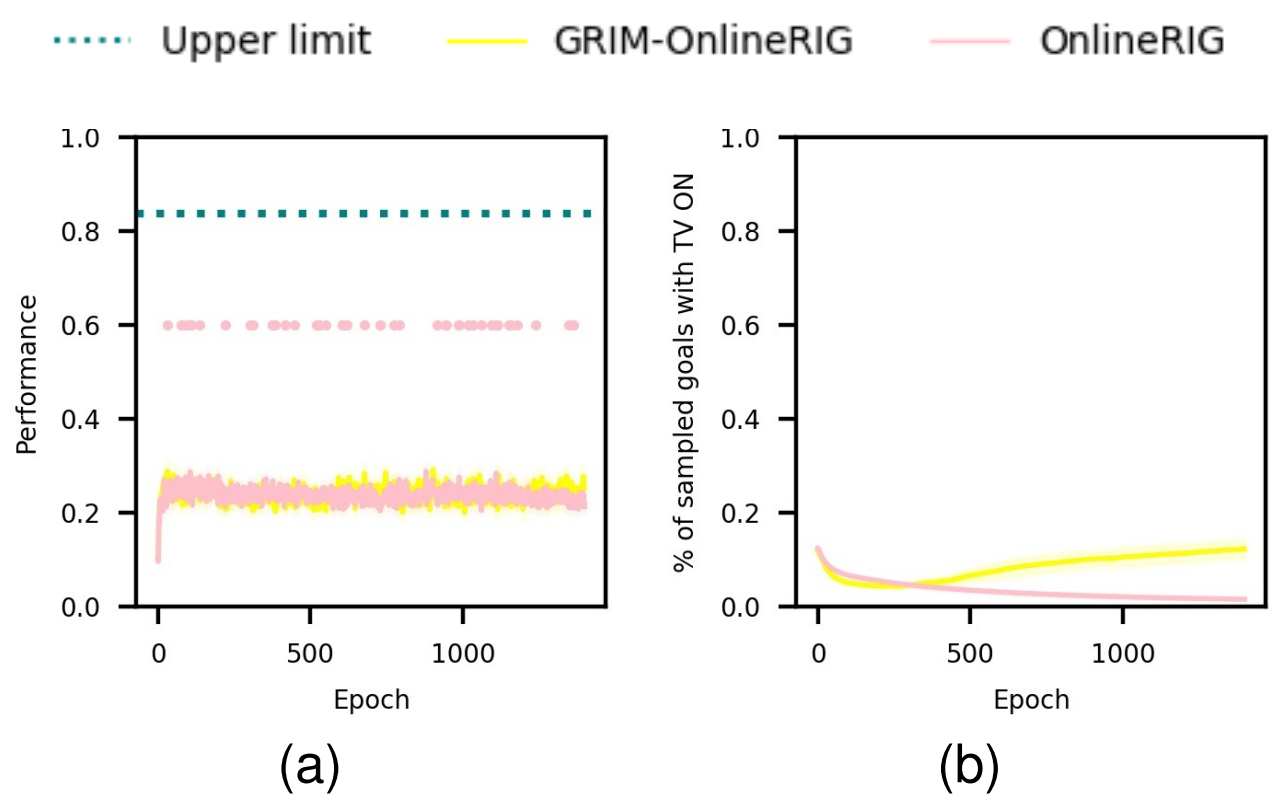

Fig. 1: The performance and the percentage of sampled goals depicting the active TV (± std. err.) The dots depict statistical significance (Welch t-test, $p = 0.05$, 10 seeds) for OnlineRIG underlying IMGEP. (a) Performance (b) % of TV ON goals

## APPENDIX D
## PRELIMINARY EXPERIMENTS IN THE 2D PLAYGROUNDRGB ENVIRONMENT

The preliminary experiments used in constructing the GRIMGEP framework were done on a simpler 2D environment called PlaygroundRGB. Since some changes were made in the architecture for extending the agent to the 3D environment, the GRIMGEP architecture used in these experiments on PlaygroundRGB environment is not the same as the one described in the main paper.

In this section, we will explain those differences, the environment, and the experiments.

### A. Changes to the GRIMGEP between the experiments in the PlaygorundRGB (2D environment) and the Explore3D

**Clustering component** Since the 2D environment does not have realistic objects, and has a simpler underlying latent space, an additional VAE was trained to create the clustering latent space on which to train the GMM i.e. the YOLO backbone, global average pooling, and the PCA were replaced with this VAE. This additional VAE is not the same VAE as the one used inside the underlying IMGEP to compute the rewards for training the policy. It is important that this VAE's latent space creates the features which are relevant for separating the regions and not on the specific details relevant for training the agent. For this purpose, we reduce the size of the VAE and its latent space. For the hyperparameters, see table II. We train this VAE online after each epoch on the data uniformly sampled from the replay buffer.

In the experiments on the PlaygroundRGB environment ten different GMM [2] models each having a different number of clusters (1, 3, ..., 19) was trained. Then we the best one was chosen by their AIC [3] score. This mechanism was inspired by the one used in [4]. The best GMM was set as the clustering function. Each epoch the process is repeated and a new GMM selected.

In the experiments on the Explore3D environment, the number of clusters was simply treated as the hyperparameter ($k$) as the procedure described above introduced unnecessary computation.

The clustering VAE hyperparameterss are shown in Table II.

| | |
|---|---|
| representation size | 3 |
| batch_size | 128 |
| beta | 1 |
| lr | 0.001 |
| Encoder | |
| kernel sizes | [5, 3] |
| num of channels | [4, 4] |
| strides | [3, 2] |
| Decoder | |
| kernel sizes | [3, 3] |
| num of channels | [4, 4] |
| strides | [2, 2] |

TABLE II: The Clustering VAE hyperparmaeters. A reduced version of the training VAE used in the underlying IMGEPs.

### B. The PlaygroundRGB environment

The *PlaygroundRGB* environment consists of three rooms through which the agent can move. The agent's actions space consists of 3 continuous actions which control the position of the gripper and can close/open /the arm. It observes the environment as an image of a top-down view of the current room.

The available rooms are depicted in Fig. 2 and the topology of the environment in Fig. 2a. The agent always starts in the Start room. All the possible goals inside this room are very easy and require only moving the gripper to the correct location. The Object room represents the *interesting* part of the environment, as it contains a movable object and is the only non-distracting part of the environment. The TV room plays the role of an action induced noisy distractor. This room contains a TV that can be turned on by closing the gripper. When the TV is turned ON, the location of the TV and the background color are randomized (a random color from a set of 5).

In short, to solve the task, exploration should guide the agent out of the Start room, away from the TV room, and into the Object room.

Since goals from the Start room are easy, we expect any goal exploration algorithm to learn goals inside this room. However, only algorithms exploring well should master goals inside the Object room. This is why, for evaluation, we construct a static test set of 25 goals from the Object room. Goals are completed if in the last state both objects are in the correct location. The performance of the agent is the average success over this evaluation set.

For evaluation, the 25 test goals from the OBJECT room were constructed by selecting 5 possible locations (center, NW, NE, SW, SE) and doing the Cartesian product of these locations for both objects. We evaluate the location of an object as correct if its L-$\infty$ distance from the goal's location is less than 0.2.

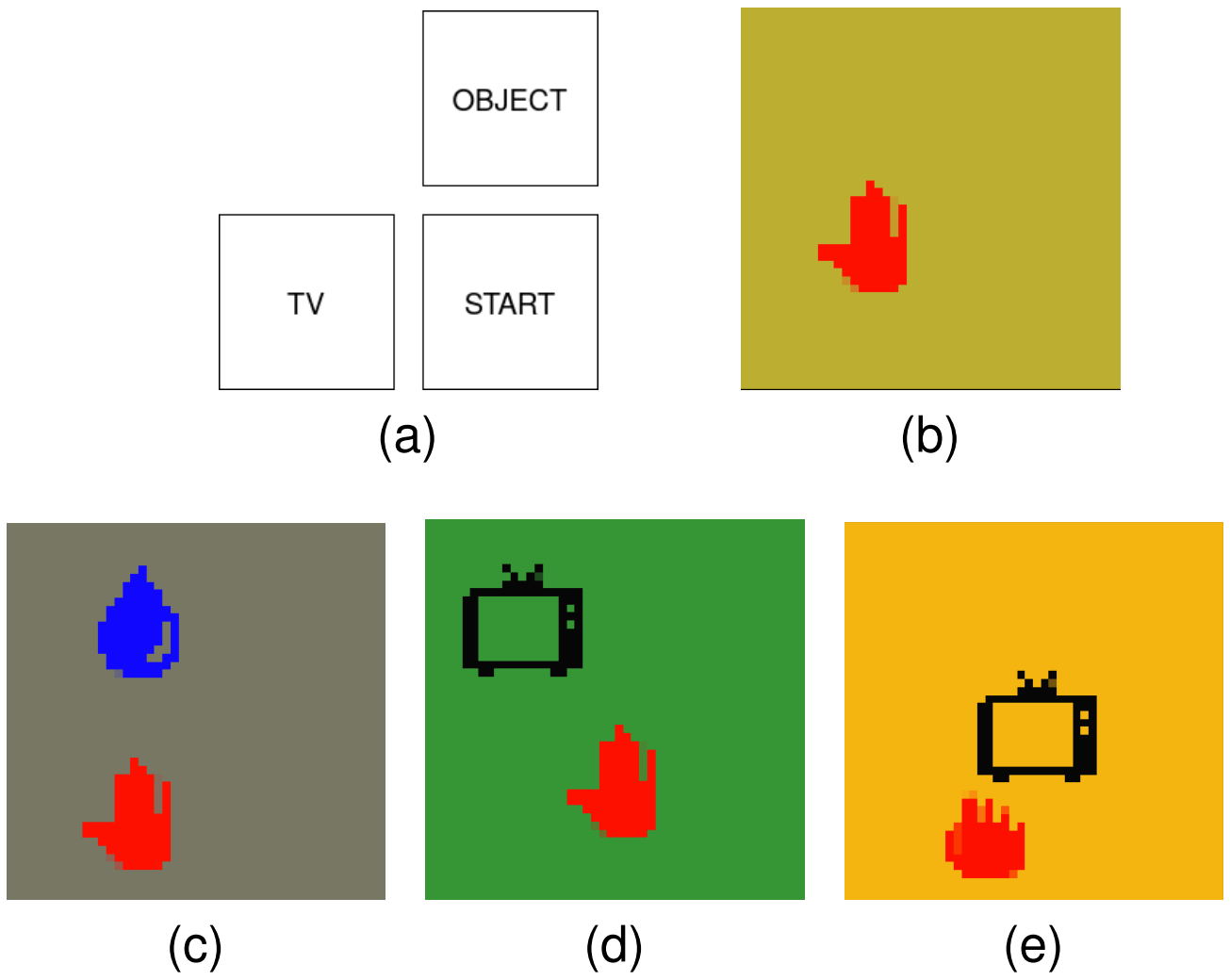

Fig. 2: Different rooms of the PlaygroundRGB environment. A preliminary 2D environment. (a) Topology (a) Start room (a) Object room (a) TV room OFF (a) TV room ON

## C. Experiments

In these experiments we address the same questions as in the Explore3D. Those questions are:

- How do current approaches behave in the presence of action-induced distractors?
- How does the GRIMGEP framework change the behavior of current approaches in the presence of noisy distractors?
- How important are the ALP estimates for the performance of the GRIMGEP framework?

We study this questions using two novelty seeking IMGEPs (Skewfit and CountBased) and one which doesn't have exploration incentives and samples goals uniformly from the history of encountered states (OnlineRIG).

**How do current approaches behave in the presence of noisy distractors?**

We can see, in Fig. 3d, that Skewfit is heavily drawn to the noisy part of the TV room resulting in very low sampling of other parts of the environment, notably the Object room. As a result of not exploring the Object room enough, the final performance diminishes. CountBased also samples a lot of goals in the TV room. However, in comparison to Skewfit, the focus is separated between both the TV-on and TV-off goals (see Fig. 3d and Fig. 3e). OnlineRIG is also not able to achieve high performances. The reason is that, due to the lack of exploration incentive, it samples goals mostly from the Start room as can be seen in Figure 3f).

Overall, this experiment demonstrates that this environment requires exploration incentives but, that these incentives should not be novelty based.

**How does the GRIMGEP framework change the behavior of current approaches in the presence of noisy distractors?**

To answer this question we wrap the Skewfit and CountBased with the GRIMGEP framework. As can be seen in figures 3c and 3d, both GRIM-Skewfit and GRIM-CountBased, focus more on the Object room and less on the TV room than their unwrapped counterparts. This results in much better performances at the end of the training (see Fig. 3a). When using OnlineRIG inside the GRIMGEP framework we can again observe a strong focus on the Object room (see Fig. 3g), but we can also see that this focus alone is not sufficient to greatly improve performance, i.e. novelty based exploration is needed.

This leads us to the conclusion that GRIMGEP successfully does two things: 1) it detects the relevant part of the goal space, and 2) successfully uses the novelty seeking exploration in this relevant region.

**How important are the ALP estimates for the performance of the GRIMGEP framework?**

We study this question by doing an ablation study on the cluster sampling technique. In the experiments in Figure 4a we test how cluster sampling based on LP (GRIM-LP-*imgep_name*) differs in comparison to uniform cluster sampling (GRIM-UNI-*imgep_name*). As shown in Fig 4a, GRIMGEPs that use LP outperform the GRIMGEPs that don't. Furthermore, we can see that, when sampling the clusters uniformly, the Object room is sampled considerably less.

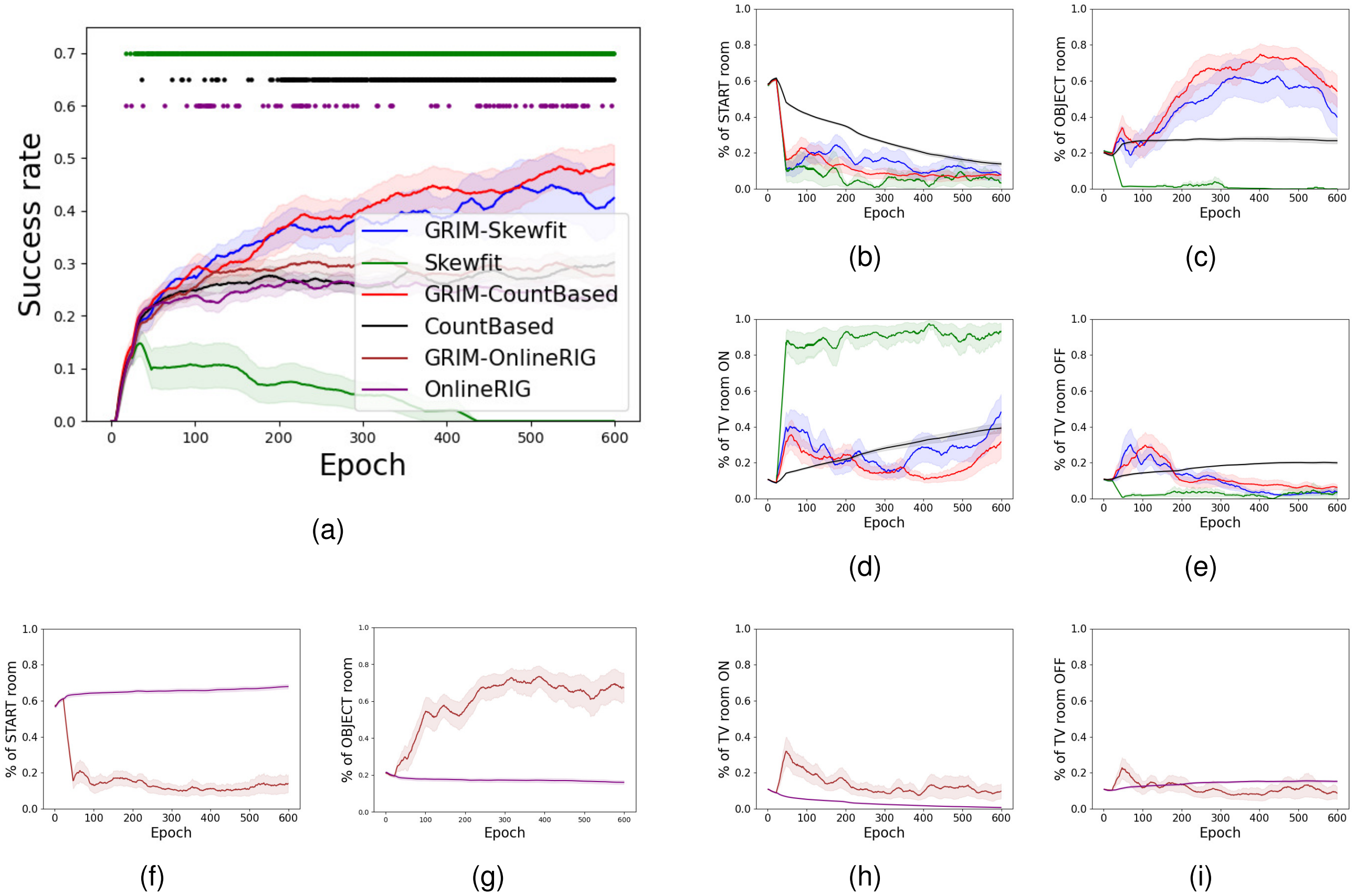

Fig. 3: Comparison of CountBased Skewfit and OnlineRIG when used alone and in combination with when used inside the GRIMGEP framework on the preliminary 2D PlaygroundRGB environment. We can see that GRIMGEP improves the performance of all Skewfit, CountBased and OnlineRIG. Furthermore, we can see that the performance of GRIM-OnlineRIG is inferior to that of GRIM-Skewfit and GRIM-CountBased. Since OnlineRIG doesn't use exploration bonuses (samples goals uniformly), this shows that GRIMGEP is able to both detect the interesting region of the goal space and reuse the novelty-based exploration of CountBased and Skewfit in this region. Ten seeds were used, and the dots depict statistically significant ($p < 0.05$, Welch's t-test) results. The shaded areas correspond to standard errors, and the bold line to the mean (smoothed over 25 epochs). (a) Success rates (b) Object room sampling for Skewfit and CountBased (c) Start room sampling for Skewfit and CountBased (d) TV room ON sampling for Skewfit and CountBased (e) TV room OFF sampling for Skewfit and CountBased (f) Object room sampling for OnlineRIG (g) Start room sampling for OnlineRIG (h) TV room ON sampling for OnlineRIG (i) TV room OFF sampling for OnlineRIG

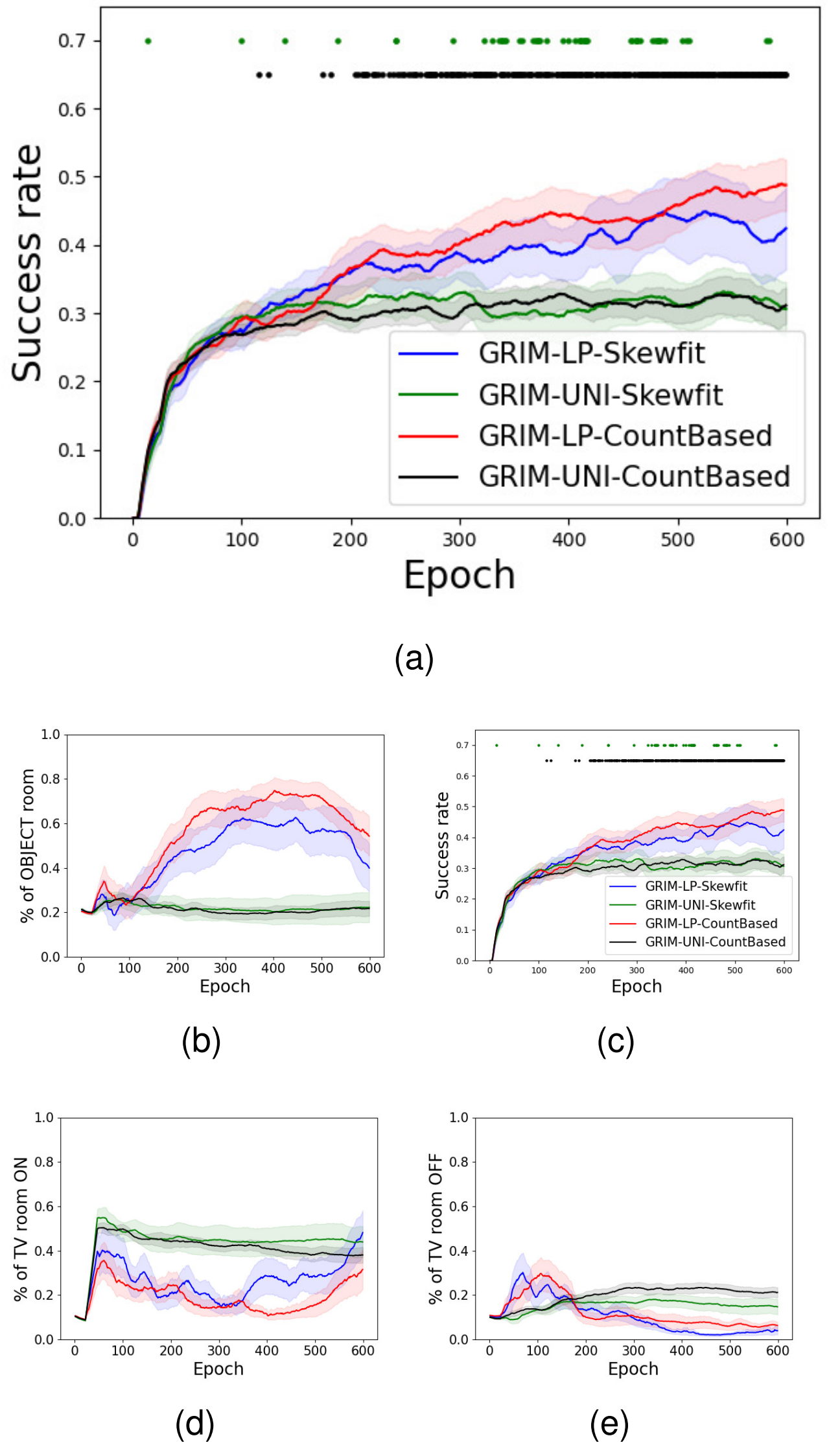

Fig. 4: Comparison of the GRIMGEP framework with the sampling of clusters according to ALP and uniformly on the preliminary 2D PlaygroundRGB environment. It is visible that GRIMGEP works better when LP is used to select the most interesting cluster. The dots depict statistically significant ($p < 0.05$, Welch's t-test) results when compared to the LP version (GRIM-LP-*imgep_name*). The shaded areas correspond to standard errors and the bold line to the mean (smoothed over 25 epochs). (a) Success rates (b) Object room (c) Start room (d) TV room ON (e) TV room OFF